\documentclass[Afour,sageh,times,doublespace]{sagej}

\usepackage{moreverb,url}

\usepackage[colorlinks,bookmarksopen,bookmarksnumbered,citecolor=red,urlcolor=red]{hyperref}

\usepackage{graphicx}
\usepackage{amsmath}
\usepackage{algorithm}
\usepackage{algorithmic}
\newcommand{\INDSTATE}[1][1]{\STATE\hspace{#1\algorithmicindent}}
\usepackage{caption}
\usepackage{subcaption}

\newcommand\BibTeX{{\rmfamily B\kern-.05em \textsc{i\kern-.025em b}\kern-.08em
T\kern-.1667em\lower.7ex\hbox{E}\kern-.125emX}}

\begin{document}

\runninghead{Kearney et al.}

\title{What's a good prediction?\\ Challenges in evaluating an agent's knowledge}
\author{Alex Kearney\affilnum{1}, and Anna J. Koop \affilnum{1}, Patrick M. Pilarski\affilnum{1,2,3,4}}

\affiliation{\affilnum{1} Department of Computing Science, \textit{University of Alberta}
\affilnum{2}Department of Medicine \textit{University of Alberta}
\affilnum{3}Alberta Machine Intelligence Institute
\affilnum{4}DeepMind}

\corrauth{Alex Kearney, hi@alexkearney.com}

\begin{abstract}
Constructing general knowledge by learning task-independent models of the world can help agents solve challenging problems. However, both constructing and evaluating such models remain an open challenge. The most common approaches to evaluating models is to assess their accuracy with respect to observable values. However, the prevailing reliance on estimator accuracy as a proxy for the usefulness of the knowledge has the potential to lead us astray. We demonstrate the conflict between accuracy and usefulness through a series of illustrative examples including both a thought experiment and an empirical example in Minecraft, using the General Value Function framework (GVF). Having identified challenges in assessing an agent’s knowledge, we propose an alternate evaluation approach that arises naturally in the online continual learning setting: we recommend evaluation by examining internal learning processes, specifically the relevance of a GVF's features to the prediction task at hand. This paper contributes a first look into evaluation of predictions through their use, an integral component of predictive knowledge which is as of yet unexplored.
\end{abstract}

\keywords{Reinforcement Learning, General Value Functions, Agent Knowledge}

\maketitle

A cornerstone of intelligence is knowledge. It is no surprise that much artificial intelligence research has been focused on designing algorithms that enable agents to construct knowledge of their world. An agent's ability to solve complex decision tasks can be improved by forming models of the world~\citep{jaderberg_reinforcement_2016,barreto_successor_2017,ha_world_2018}. The term model is sometimes restricted to estimating the probability of state transitions. In this paper, we take a broader view of what counts as a model, including predictions that forecast future input values an agent might experience. In this sense, agents construct knowledge of their world by learning to model and forecast aspects of the environment it inhabits.

The benefits of constructing knowledge by forecasting inputs are evident in computational reinforcement learning~\cite{sutton_reinforcement_1998}, where an agent must learn to act optimally in order to maximise some expected cumulative future reward. Instead of finding the optimal policy directly, agents often learn the expected reward, or \textit{value}, of states in their environment. By learning the value of a state, it becomes easier to determine what the optimal actions are.


Value functions are deeply related to the problem of control, and the distinction between the main task (finding the optimal policy) and model (estimating the value of a state) is subtle. However, modelling the environment need not end with estimating the value of states: modelling other aspects of the environment can also support decision-making~\cite{comanici_knowledge_2018,koop_investigating_2008,jaderberg_reinforcement_2016,modayil_multi-timescale_2014,edwards_machine_2016,white_developing_2015}. For instance, it may be useful for an agent to estimate how different inputs change in response to its behaviour~\cite{jaderberg_reinforcement_2016, sherstan_work_2020}: how an agent can control what it observes through its actions. These models of the world that are independent of a particular task or goal an agent is trying to achieve can be used flexibly across different problems, including new and unseen tasks~\cite{sherstan_accelerating_2018,barreto_successor_2017}.

Learning models independent of the main task not only supports agents in solving complex problems, it also forms general knowledge of the world that can be applied to new and unseen problems. How well an agent has acquired knowledge is often measured using quantitative metrics: e.g., by directly measuring accuracy of a model's estimate~\cite{pilarski_steps_2016,modayil_multi-timescale_2014,sutton_horde_2011}, or by examining reward received by an agent on the main task~\cite{jaderberg_reinforcement_2016, schlegel_general_2018}. Systems with better quantitative outcomes are  believed to better encode knowledge on a particular task.

 As the main contribution of this paper, \textbf{we argue that evaluating knowledge is not the same as evaluating task performance: there are new challenges that need to be addressed}. In particular, a model with higher estimated accuracy does not imply that the model supports learning to solve the main problem, or task.
 
We introduce this distinction this by constructing two examples: first, where traditional evaluation techniques lead to poor model choices; second, where poor model choices have down-stream consequences when used to inform decision-making. Finally, we posit that by examining internal learning processes, we can begin to evaluate agent knowledge, and show an example of how this may indeed be possible.

\begin{figure*}[htbp!]
    \centering
    \begin{subfigure}[b]{0.3\textwidth}
        \centering
        \includegraphics[width=\linewidth, height=1.6in, keepaspectratio]{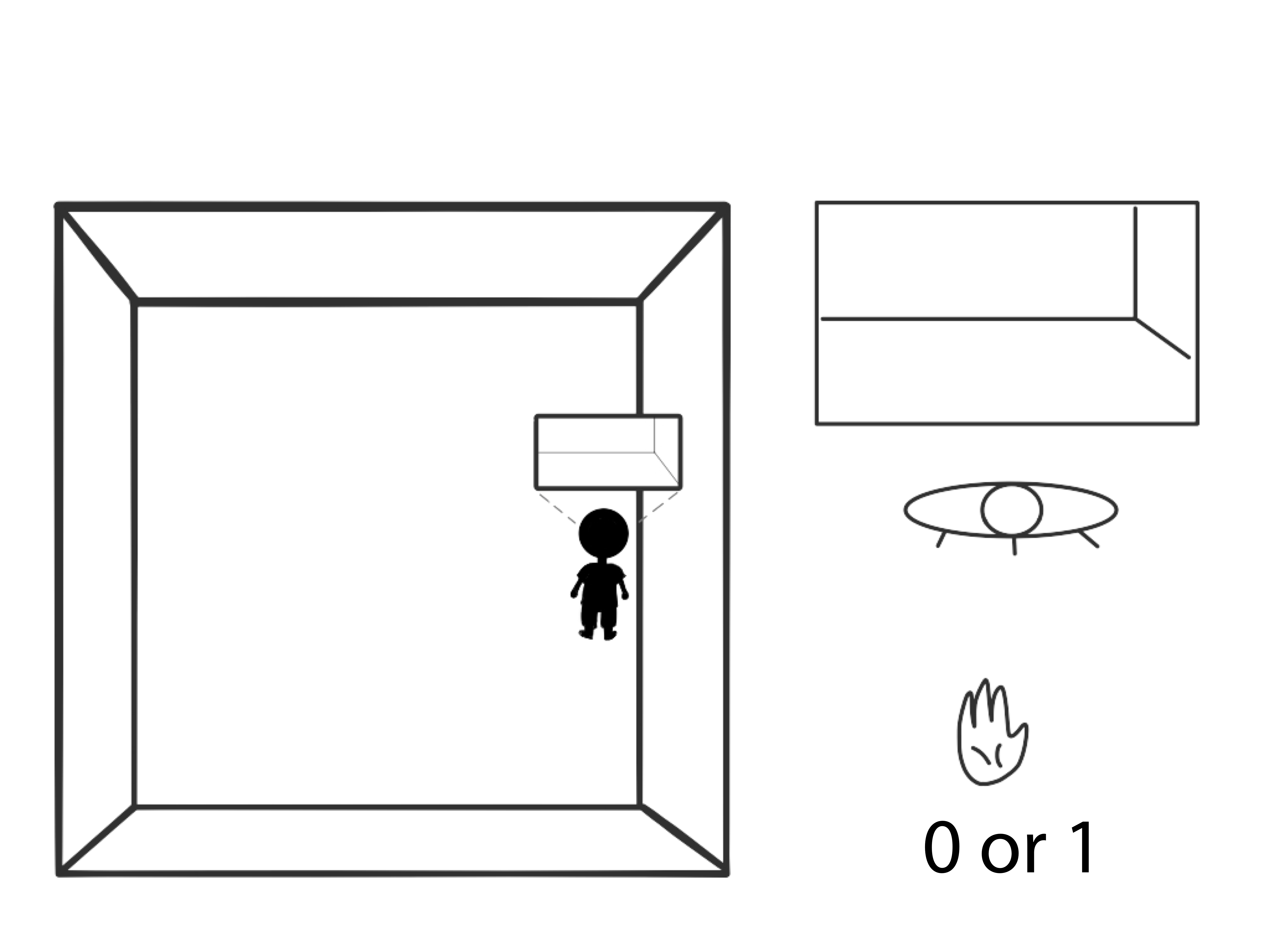}
\caption{Often an agent cannot observe the true state of the environment; e.g., an agent in a room may only observe what it can see in front of itself and whether the agent bumped into something.}\label{prediction}
    \end{subfigure}%
  \hspace{1em}%
    \begin{subfigure}[b]{0.3\textwidth}
        \centering
        \includegraphics[height=1.6in, keepaspectratio]{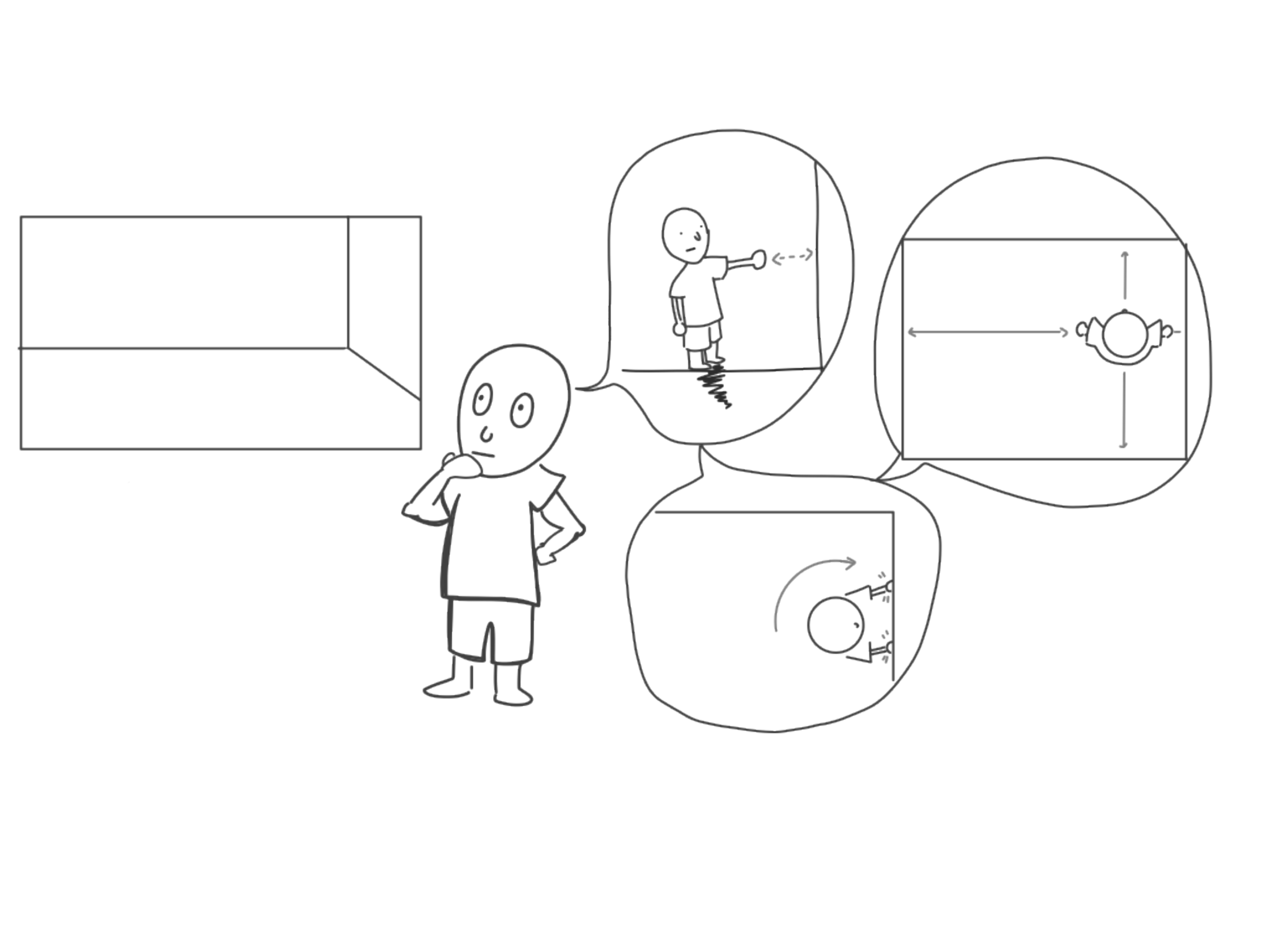}
\caption{Using limited sight and touch sensation, we can phrase basic spatial awareness as making predictions about moving around the room: e.g., ``can I touch something in front of me?'', or ``how far is the nearest wall to my left''?}\label{error}
    \end{subfigure}%
    \hspace{1em}%
    \begin{subfigure}[b]{0.3\textwidth}
    \centering
          \includegraphics[width=\linewidth,height=1.6in,keepaspectratio]{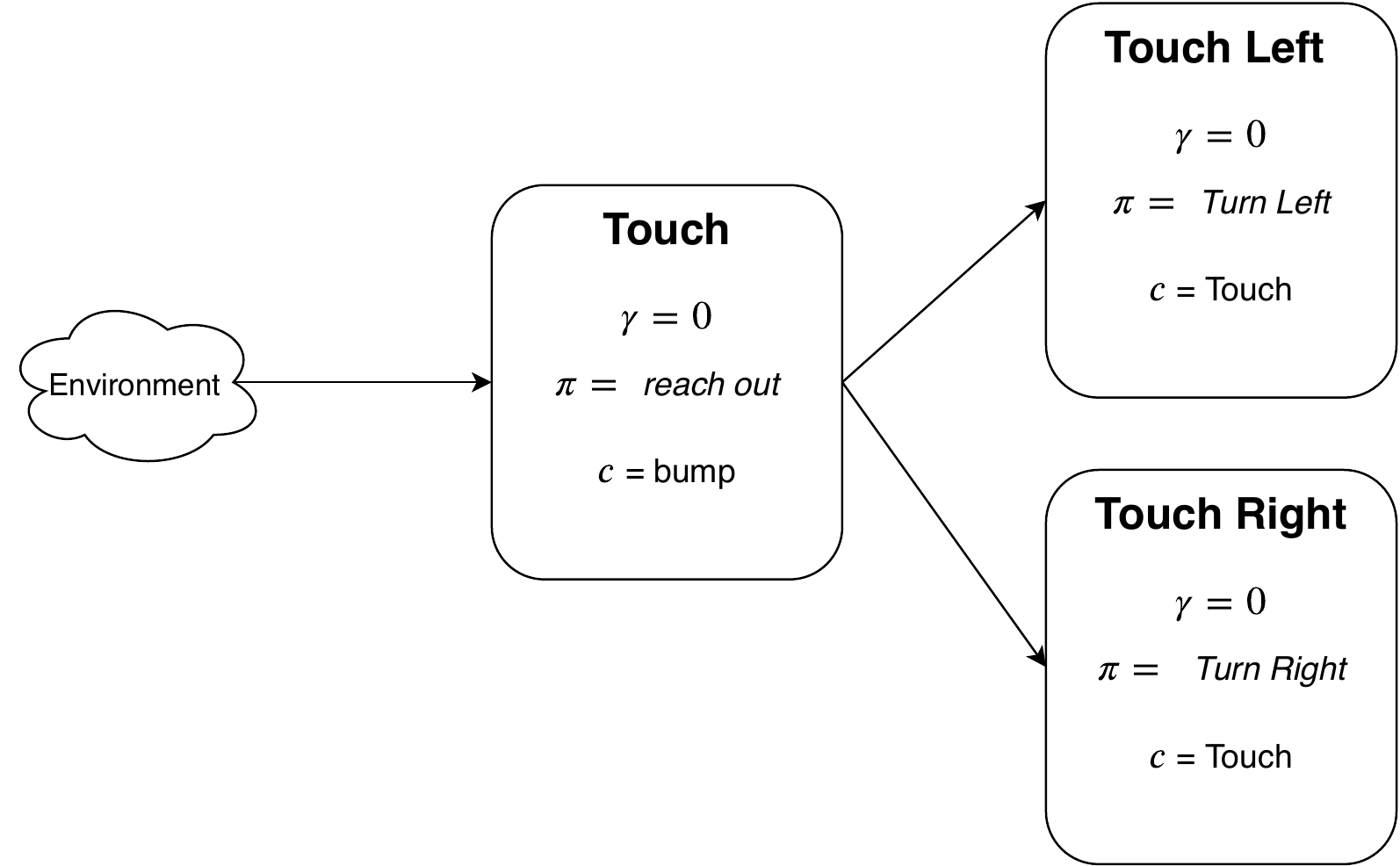}
    \caption{
    A prediction about bumping is used to construct a \texttt{touch} prediction, the output of which is used as the target for the \texttt{touch-left} and \texttt{touch-right} predictions. Adapted from Ring~\cite{ring_representing_2016}.
    }\label{layering_gvfs}
    \end{subfigure}
    \caption{Using the limited senses available to the agent, it must construct an abstraction such that it can understand a world it can never completely see. One way of constructing an agent's knowledge of the world is by predicting what would happen if the agent behaved a certain way. }\label{motivation}
\end{figure*}

\section{Understanding the world through general value functions}

Although our arguments  apply broadly to evaluating machine learning methods, we ground our arguments in a single learning problem of interest: learning predictions as an agent interacts with its world. Predictions play an important in the construction of knowledge both machines and also biological intelligence. Humans and animals continually make many predictions about their sensations~\cite{rao_predictive_1999,pezzulo_grounding_2011,pezzulo_human_2013,wolpert_internal_1995,clark_whatever_2013,gilbert_stumbling_2009,noe_action_2004}. With this in mind, we use predictions to for discussion of analysing knowledge in machines.

General Value Functions are a way for machines to learn and make predictions incrementally and online, as an agent interacts with the environment~\cite{sutton_horde_2011}. GVFs are entirely self-supervised and can be learned independent of the task an agent is undertaking through off-policy learning~\cite{sutton_horde_2011}. In this paper we use GVFs as a computational tool to enable us to clearly make our arguments, although our arguments are independent of GVFs themselves and broadly applicable situations where models are evaluated independent of their use.


General Value Functions estimate the value of a signal in a sequential decision making process. On each time-step $t$ an agent observes inputs $o_t$ from the environment and takes an action $a_t$ which results in a change in the environment, and thus a new observation $o_{t+1}$. GVFs~\footnote{General Value Functions are a generalisation of ordinary value functions: a central component of computational reinforcement learning. Value functions are used to estimate the value of a given state of the environment. General Value Functions are a generalization of traditional Value Functions to not just reward, but any signal accessible to the agent through its experience.} estimate the future accumulation of a \textit{cumulant} $c$, where $c$ is some signal of interest available to the agent through its subjective stream of experience. In the simplest case, this might be the accumulation of some element of an agent's observation $c \in o$. The discounted sum of $c$, is called the \textit{return}, and is defined over discrete time-steps $t$  as $G_t = \mathbb{E}_\pi[ \sum^\infty_{k=0}(\prod^{k}_{j=1}(\gamma_{t+j}))C_{t+k+1}]$---the expectation of how a signal will accumulate over time.

When humans interact with the environment, they construct models of the world by constantly forecasting and anticipating what will happen next~\cite{gilbert_stumbling_2009,wolpert_internal_1995, rao_predictive_1999}. Similarly, an agent can build up self-supervised models that describe the environment predictive questions such as ``If I do this, I expect that'' with~\cite{sutton_horde_2011,comanici_knowledge_2018,ring_representing_2016}. An agent can achieve greater complexity by beginning with simple, primitive predictions about future features, and interrelating them---making forecasts of forecasts. Such primitive predictions can inform more complex predictions in two ways: one prediction may be used as an input in another; or, one prediction may be used as a cumulant $c$ of another prediction---one GVF's prediction may be what another GVF is predicting. By interrelating predictions we are able to express abstract concepts that extend beyond the immediate observation stream.

Predictions as knowledge are constructed by starting with low-level immediate predictions about sensation. For example, an agent may begin to build a model of spatial awareness by predicting whether there is something in front of it: if the agent reaches out, would it be able to touch something? This simple primitive prediction could be used to inform more abstract models: e.g.\ if the agent were to turn left or right, would there be something next to it? How far away is the nearest wall? By interrelating predictive models, we can express more abstract, conceptual aspects of the environment~\cite{ring_child_1997,ring_representing_2016,koop_investigating_2008,sutton_between_1998, comanici_knowledge_2018, singh2005intrinsically} (in this case, spatial awareness) in a self-supervised way.


We can estimate GVFs using Temporal-difference (TD) learning~\cite{sutton_learning_1988}. In TD learning we estimate a value-function $v$ such that $v(\phi(o_t)) \approx \mathbb{E}_\pi [G_t | o_t]$: we learn a function that estimates the return at a given time-step given the agent's observations. On each time-step the agent receives a vector of observations $o \in \mathbb{R}^m$. A function approximator $\phi : o \rightarrow \mathbb{R}^n$---such as a neural net, Kanerva coder, or tile coder---encodes the observations into a \textit{feature vector}. The estimate for each time-step  $v(\phi(o_t))$ function of learned weights $w\in \mathbb{R}^n$, and the current feature vector---$v(o_t) = w^\top\phi(o_t)$.

We call the parameters of the learning methods \textit{learning parameters}. Learning parameters change how the value function is approximated, but do not change what the value function is about. Learning parameters include the step-size\footnote{Also known as the \emph{learning rate}.} $\alpha$ which scales updates to the weights, the eligibility trace decay $\lambda$ and the function-approximator $\phi$ used to construct state.

\section{A broken clock is right twice a day}

It is impossible to know everything about the world. Certainly, an agent cannot predict everything about its world. One challenge for constructing models of the world is deciding of all the predictions an agent could learn to make, which subset can inform decision-making best. That is, an agent must choose from all the possible predictions which it could make, the subset of predictions that will help it achieve its goals. Not all predictions are created equally: two approximate GVFs may have the same question parameters---$\gamma$, $\pi$, and $c$---and yet produce very different estimates. This is due to two factors: the learning parameters chosen, and the distribution of experience trained on. Feature construction, the step-size parameter, and especially nature of the experience contribute to the how well an estimator can be learned. To be able to compare estimators, we must have some metric or means of evaluating them. In this section we construct and example where traditional evaluation techniques lead to poor model choices.

We cannot compare GVFs to the true expected return of their cumulant $c$: we do not have access to the true return from the stream of data available to an agent. Instead, we often assess a GVF's accuracy based on an estimate of the true return, the \textit{empirical return error}: the difference between the current estimate $v(\phi(o_t))$ with an approximation of the true return \cite{pilarski_dynamic_2012,edwards_machine_2016,gunther_intelligent_2016}. The return is estimated by maintaining a buffer of length $b$  of previous cumulants $c$, such that $ \tilde{G_t} = \sum^\textit{b}_{k=0}(\prod^{k}_{j=1}(\gamma_{t+j}))C_{t+k+1})$. We may then construct an error for time-step $t$ given the agent's experience by $\tilde{G_t} - V_{t}(\phi(o_{t})$.  The empirical return error is not objective. Note, it depends on what the agent happens to experience---it can only express the error for observations represented in the buffer. It does not capture error for all possible observations or states of the world. 

In simple Markov Reward Processes this may not be an issue: maintaining a large enough buffer $b$ will yield an error relatively unbiased over states. However, in many domains of interest, this is not possible: i.e., in robotics the state-space is often so immense that maintaining a buffer of observations would be a time-intensive and impractical demand.  Instead, applications often settle for an empirical return error that covers only a portion of the state-space~\cite{edwards_machine_2016,gunther_intelligent_2016,gunther_examining_2020,pilarski_steps_2016}. In doing so, some states are inherently prioritised over others, as they are gerrymandered into two categories: the portions of state-space that are evaluated, and the portions that are not. When evaluating methods in this way, it is implicit that some of the states are privileged over others: that error matters more in one set of states over another~\cite{sutton2009fast}.

\begin{figure}[tp!]
    \centering
    \includegraphics[height=2.1in, keepaspectratio]{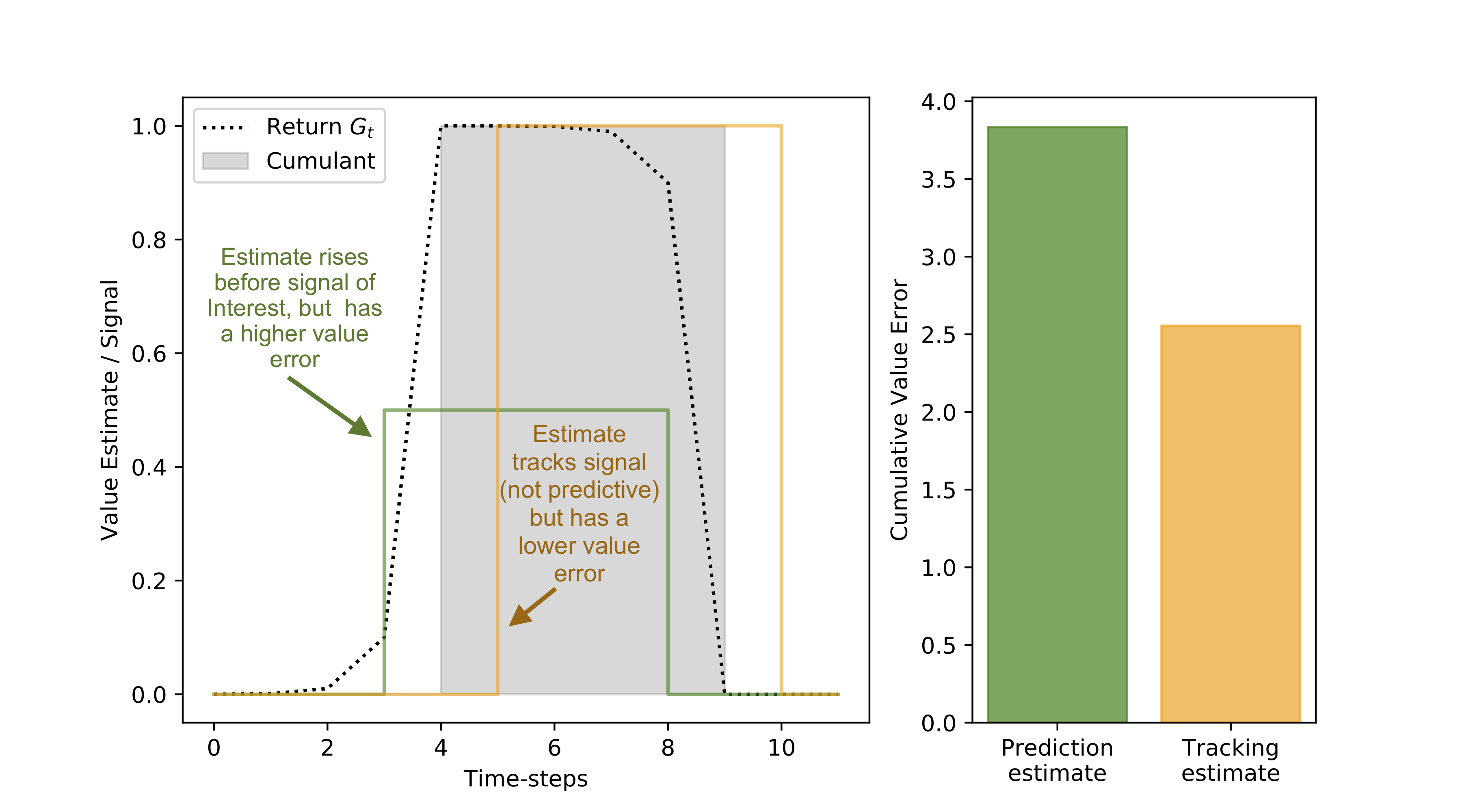}
    \caption{Two estimates of the same signal: one in green and one in orange.}
    \label{comparison}
\end{figure}

As an example of how empirical error can be gerrymandered by state we present two hypothetical estimators of the same value-function in figure~\ref{comparison}. A binary square-pulse is the cumulant $c$ for which two functions estimate the discounted return. The dotted line is the scaled return $G_t$ of the cumulant $c$ with a discount factor of $\gamma = 0.3$ that is being estimated. An ideal prediction that matches the return $G$ of the signal, should rise before the signal of interest $c$ rises, and fall before the pulse returns to 0. This value estimate is predictive, it rises before the signal of interest rises, and falls before the signal of interest falls---it precedes the signal of interest. The tracking estimator's estimate is whatever the cumulant was at the previous time-step. The tracking estimator is not predictive: it rises and falls after the signal of interest. In spite of this, the non-predictive estimator has the lower empirical return error.

If we were evaluating the two predictions and choosing between these two estimators using prediction error alone, we would be led to believe that the tracking estimator is superior to the predictive estimator: it has a lower cumulative error. This becomes an issue when these estimates are intended to inform decision-making. For instance, if an agent is predicting a collision, identifying the collision has occurred after the fact is not useful in supporting decision-making.

While this example is contrived, there are many situations in which we would want to make such a prediction; being able to detect the onset of events is often useful in decision-making~\cite{modayil_prediction_2014,schlegel_general_2018,ring_representing_2016}. For example, in the previous section, we worked out an example where an agent built a sense of spatial awareness (Figure~\ref{motivation}) by predicting whether it could touch something in front of itself; In the spatial awareness example, touch is a binary signal that rises and falls, similar to this simple synthetic example. Such predictions are not made in a vacuum: the motivation for learning models is to use them to inform decision-making.

\section{Downstream consequences of poor evaluation}\label{section:rupee}

\begin{figure*}
\centering
    \begin{subfigure}[b]{0.45\textwidth}
    \centering
    \includegraphics[width=\linewidth, keepaspectratio]{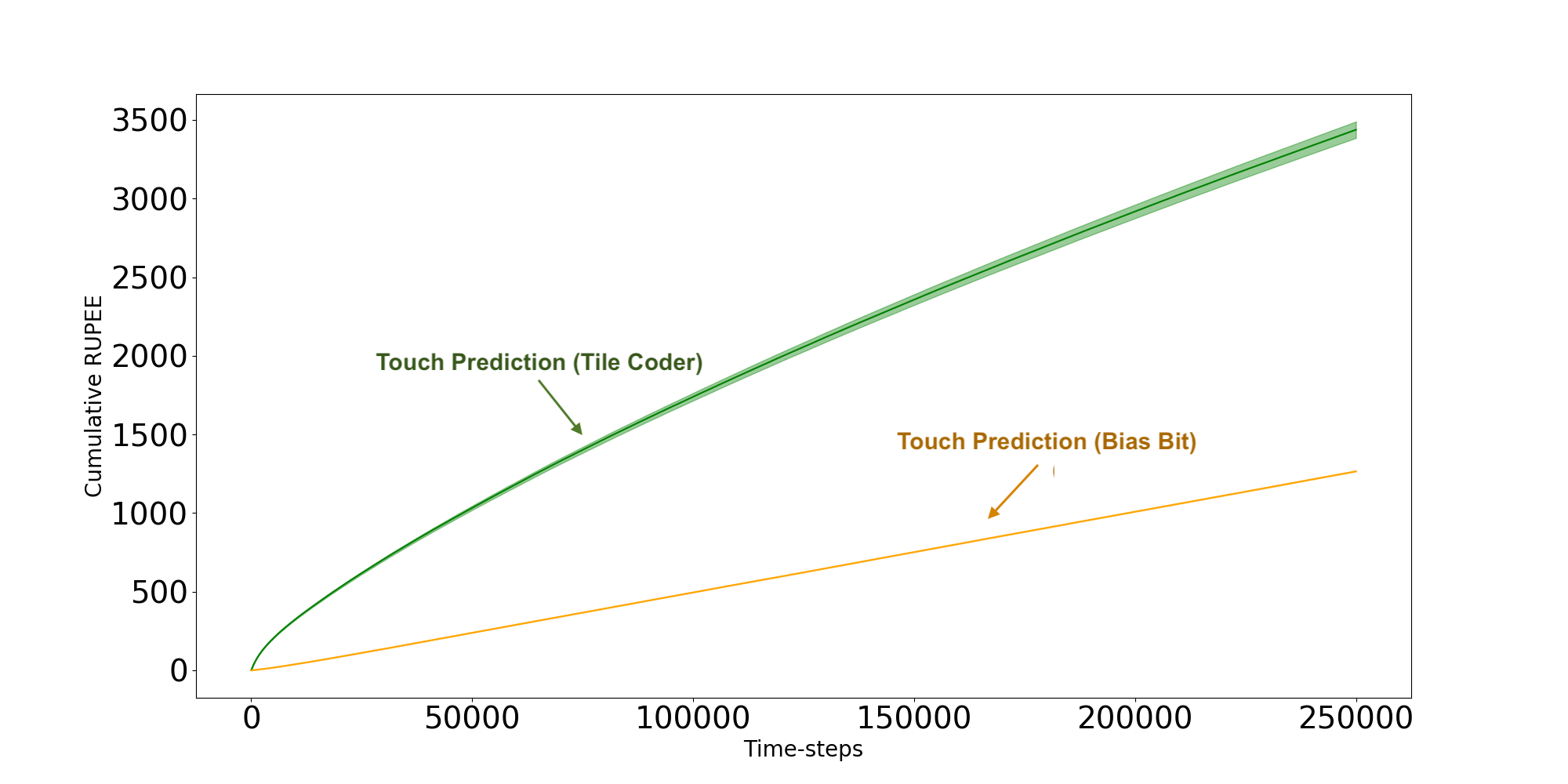}
    \caption{Cumulative RUPEE for tile-coded \texttt{touch} estimate (green) and bias-bit \texttt{touch} estimate (orange).The tracking estimate accumulates error at a slower ratethan the anticipatory prediction. Evaluating based on RUPEE alone, we would be led to believe that the tracking model is best, despite leading to catestrophic prediction error when used to inform \texttt{touch-left} and \texttt{touch-right} (c.f. Figure~\ref{vingette}). The anticipatory touch estimate has a greater accumulation of error throughout the experiment despite being a better estimator for informing \texttt{touch-left} and \texttt{touch-right} predictions. \\ }
    \label{rupee_first}
    \end{subfigure}%
     \hspace{1em}%
  	\begin{subfigure}[b]{0.45\textwidth}
    \centering
    \includegraphics[width=\linewidth, keepaspectratio]{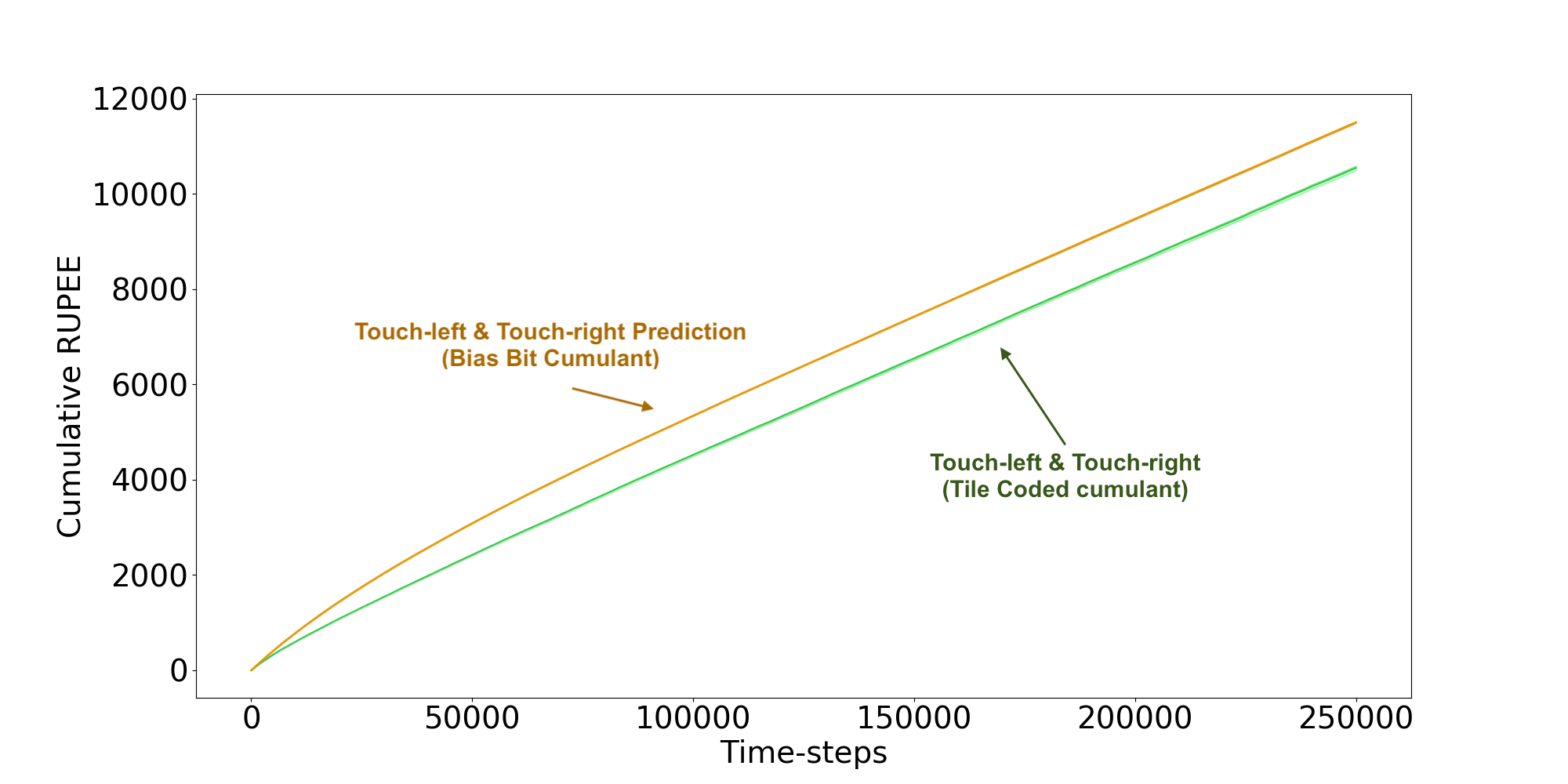}
    \caption{Cumulative RUPEE for \texttt{touch-left} and \texttt{touch-right} estimates which use as a cumulant the tile-coded (green)  and bias bit  (orange) \texttt{touch} estimate. Estimates dependent on the tracking GVF for learning have a greater cumulative error than the GVFs dependent on the Tile Coder GVF. Error as accumulated at roughly the same rate as the anticipatory GVFs, making it challenging to distinguish which of the prediction is better, despite wildly different outcomes when comparing prediction to ground-truth (c.f. Figure~\ref{vingette}). The error of the lower-order models does not always determine their effectiveness in informing further learning.}
    \label{rupee_second}
    \end{subfigure}
     \caption{Cumulative Recent Unsigned Projected Error Estimate (RUPEE over 250,000 time-steps for the `touch-left' and `touch-right' predictions averaged over 30 independent trials.}
     \label{rupee}
\end{figure*}

With a simple example, we have demonstrated how accuracy can be misleading in differentiating between forecasts. We now discuss how dependence on accuracy negatively impacts down-stream learning processes and critically undermines representation learning.

A motivation of constructing general knowledge using GVFs is the ability to build modular, and hierarchical forecasts about the world. This is achieved by 1) using an estimate as an input feature when making a higher-order GVF, or 2) using a learned estimate as a cumulant for another GVF. In this section, we demonstrate that poor evaluation in lower-order GVFs has consequences for the performance of higher-order GVFs. In order to demonstrate these challenges in evaluation, we turn our attention to the off-policy prediction setting.

An off-policy error metric which can be calculated incrementally online, is RUPEE:\@ the Recent Unsigned Projected Error Estimate~\cite{white_developing_2015}. RUPEE estimates the mean squared projected Bellman error of a single GVF\footnote{See~\cite{white_developing_2015} for an explanation of RUPEE on pages 119-122.}. Intuitively, RUPEE is an estimate of learning progress with respect to the input features used by the agent in learning. While RUPEE does not imply prediction accuracy, RUPEE provides a computationally efficient way to determine when a forecast learned off-policy is approaching its best estimate~\cite{white_developing_2015}.

RUPEE requires an additional parameter $\beta_{0}>0$ which specifies a decay rate for the exponential moving average of both $\tau$ and $\overline{\delta e}$. A higher $\beta$ value results in a longer horizon for the moving average. Where $e$ are the forecast's eligibility traces, $\delta$ is the TD error, and $h$ is the same as the update in GTD ($\lambda$); RUPEE is estimated as follows:

\[
\tau \gets (1-\beta_{0})\tau + \beta_{0}
\]

\[
  \beta \gets \frac{\beta_{0}}{\tau}
\]

\[
  \overline{\delta e} \gets (1-\beta) \overline{\delta e} + \beta \delta e
\]

\[
  \text{RUPEE} \gets \sqrt{|\hat{h}^{\top} \overline{\delta e ^{\beta}}|}
\]

As was the case when evaluating on-policy predictions via empirical return error, by estimating off-policy learning progress using RUPEE, we are unable to differentiate between useful and useless estimators. We demonstrate this evaluation conundrum's down-stream effects of imprecise evaluation of predictions: whether the agent could touch something if it extends its hand, and whether the agent could touch something if it turned left or right (as introduced in Figure~\ref{motivation}). These predictions are useful building-blocks that can inform much more complex predictions that express abstract aspects of the world---i.e., basic navigation and spacial awareness~\cite{ring_representing_2016}. In order to get to these higher-order, abstractions, we must first be able to get these simple, primary predictions right.

\begin{figure}[!t]
    \centering
    \begin{subfigure}[b]{0.3\textwidth}
        \includegraphics[width=\linewidth, height=2in, keepaspectratio]{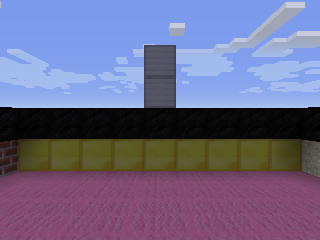}
        \caption{The visual input the agent totaling 320$\times$480 pixels.}\label{fig:gull}
    \end{subfigure} 
    
    \begin{subfigure}[b]{0.3\textwidth}
        \includegraphics[width=\linewidth, height=2in, keepaspectratio]{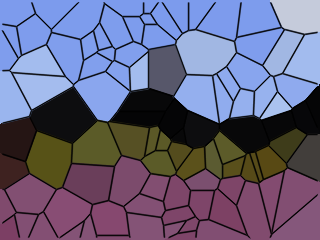}
        \caption{Visualization of the image subsampled to 100 random pixels}\label{fig:subsample}
    \end{subfigure} 
    
    \caption{A visual representation of our agent approximating the visual input by subsampling 100 random pixels.}\label{fig:voronoi}
\end{figure}

These predictions are made in a MineCraft~\cite{johnson2016malmo} grid-world that reflects the spatial awareness task we previously introduced (Figure~\ref{motivation})\footnote{This example is a simplification of the thought experiment introduced in Ring~\cite{ring_representing_2016}}. The world is a square pen which is 30 $\times$ 30 and two blocks high. The mid-section of each wall has a silver column, and the base of each wall is a unique colour. On every time-step, the agent receives observations $o_t$ which contain: 1) the pixel input from the environment (Figure~\ref{fig:gull}), and 2) whether or not the agent is touching something.

Similar to the previous synthetic example, we have two sets of value functions: one that predicts, and one that tracks. We construct two GVF networks that are specified with the same question parameters, but differ in answer parameters used. Both sets of GVFs are approximating the same values; however, the way they learn their approximation differs. One \texttt{touch} prediction uses a Tile Coder~\cite{sutton_reinforcement_1998,sherstov2005function} as a function approximator, and the tracking GVF uses only a single bias bit as a representation. We choose this representation, as it is clear that a bias bit is insufficient to inform any of the chosen predictions: we cannot predict whether the agent can touch a wall using a single bit to represent our MineCraft world.

\begin{figure}[htbp!]
\centering
\vspace{-0.5cm}
 	\begin{subfigure}[b]{\linewidth}
    	\centering
    	\includegraphics[width=\linewidth,
    	height=3in,
    	keepaspectratio]{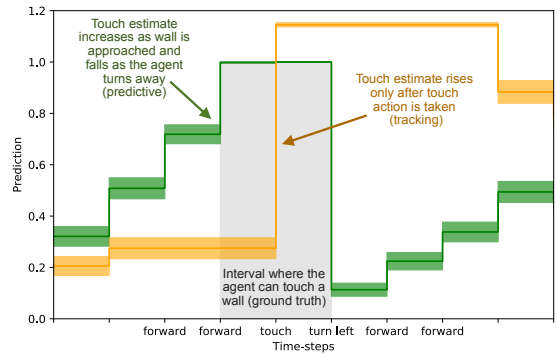}
    	\vspace{-0.7cm}
    	\caption{Tile-coded \texttt{touch} estimate (green) and bias-bit \texttt{touch} estimate (orange) \\}\label{touch_predict}
    \end{subfigure}
     \begin{subfigure}[b]{\linewidth}
     \vspace{-0.7cm}
    	\centering
    	\includegraphics[width=\linewidth, height=3in, keepaspectratio]{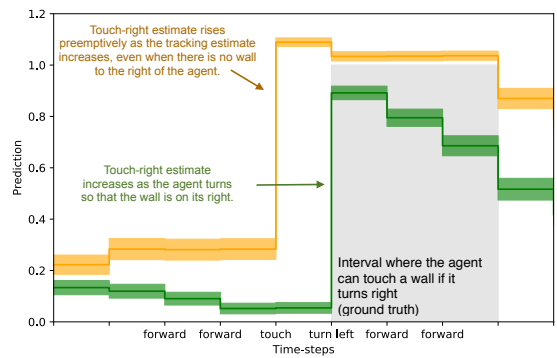}
    	\vspace{-0.7cm}
    	\caption{\texttt{touch-right} estimates which use as a cumulant the tile-coded (green)  and bias bit  (orange) \texttt{touch} estimate.}
    \label{right_predict}
    \end{subfigure}
	\begin{subfigure}[b]{\linewidth}
	\vspace{-0.7cm}
    	\centering
    	\includegraphics[width=\linewidth, height=3in, keepaspectratio]{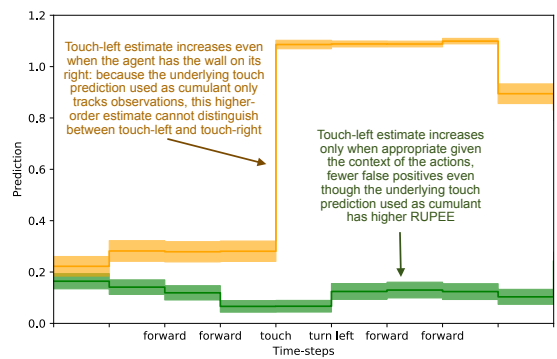}
    	\caption{\texttt{touch-left} estimates which use as a cumulant the tile-coded (green)  and bias bit  (orange) \texttt{touch} estimate.}\label{fakeout}
    \end{subfigure}
    \caption{Each sub-figure depicts estimates of each of the GVFs in our networks for 150 examples of the agent approaching a wall and then turning left. 5 examples of the trajectory are drawn from 30 independent trials.}
    \label{vingette}
\end{figure}

This experimental setup directly parallels our on-policy synthetic example in a more complex environment. As was the case in the previous thought experiment, by comparing the two \texttt{touch} predictions based on their error (Figure~\ref{rupee_first}), we would be lead to conclude that the bias bit GVF is superior to the tile-coded GVF---we would conclude that the prediction that does not predict is superior. When we examine the actual predictions made by each GVF, we see that the predictive estimate with a greater RUPEE more closely anticipates the signal of interest (Figure~\ref{touch_predict}). The reason why the bias bit prediction is poor is because it tracks. An architect designing a system understands this prediction is poor because it is redundant: the immediate sensation of touch tells us whether or not an agent \emph{is} touching something. The intent of the prediction is to compactly express whether or not an agent \emph{can} touch a wall without needing to engage in the behaviour. When the agent does touch a wall, the prediction is updated and stored in the weights of the GVF. Only when the agent is touching a wall will the bias bit GVF predict that it can touch a wall. By looking at RUPEE alone, we miss this critical shortcoming.

These predictions are not learned in a vacuum: the purpose of making the \texttt{touch} prediction, is to enable the higher-order predictions to be learned. In systems that use GVFs to construct an agent's knowledge of the world, these predictions are intended to inform further learning processes: either other value functions that describe more abstract aspects of the world, or the behaviours an agent uses to accomplish its goals. Low RUPEE or low return error in an estimator does not necessarily equate to more useful predictions for these further decision-making purposes. The challenges of differentiating between a good and bad \texttt{touch} prediction have an impact which extends beyond the single prediction and influences the \texttt{touch-left} and \texttt{touch-right predictions}.

We want not only an accurate \texttt{touch} prediction, but one which is capable of informing \texttt{Touch Left} and \texttt{Touch Right} predictions. In Figure \ref{vingette}, we display the RUPEE of \texttt{Touch Left} and \texttt{Touch Right}. There are two sets of these predictions: the first, using the bias bit GVF's prediction as its cumulant; the second, using the tile-coded GVF as its cumulant. In this layer, the GVFs all share the same function approximator: they both use sufficient representations to learn a reasonable estimate. In this case, a random sub-sampling of the pixel input, binary touch signal, and \texttt{touch} prediction are all tiled together to construct the state for each GVF. The only differentiating factor is which cumulant is used: the prediction from either the tracking \texttt{touch} GVF, or the anticipatory \texttt{touch} GVF.
 
When we examined the first layer's \texttt{Touch} predictions, the tracking GVF seemed superior based on RUPEE. When we examine the RUPEE of the second set of predictions (Figure \ref{rupee_second}), we catch a glimpse of the down-stream effects of this misunderstanding. Although only slight, the GVFs dependent on the tracking \texttt{Touch} prediction have a higher RUPEE than those using the predictive \texttt{Touch} GVF. This point is brought into focus when we examine the predictions made by each \texttt{touch-left} and \texttt{touch-right} prediction (Figures~\ref{right_predict} and~\ref{fakeout}). When we examine average trajectories where the agent approaches a wall and turns left, the \texttt{touch-right} prediction using the tracking \texttt{touch} GVF as a cumulant (Figure~\ref{right_predict}, in orange) rises and falls with its underlying GVF. The \texttt{touch-right} prediction with a tracking cumulant predicts wall even before turning such that the wall is to its right, while the \texttt{touch-right} prediction with a predictive cumulant is able to better match the ground-truth. This disparity is further exacerbated in Figure~\ref{fakeout}, where we see that the \texttt{touch-left} prediction dependent on the tracking \texttt{touch} GVF as a cumulant incorrectly anticipates a wall is on its left, even as it turns away from it. Through examining the error---the metric used to inform predictive knowledge architectures---we miss this. The use of a prediction tells us more about the quality of that prediction than error alone. By using a poor underlying \texttt{touch} prediction, the higher-order GVFs become un-learnable.

We demonstrated that poor behaviour of estimates can be hidden by commonly used error metrics. This kind of inquiry into the structure of predictions is not easily automated: it relies on inspection by system designers. Moreover, these precise comparison are limited to simple domains. The room our agent inhabits is so simple that we can acquire the ground-truth in order to examine the predictions as is done in Figure~\ref{vingette}. In many domains of interest, this ease of comparison is simply impossible. Each of these factors further frustrates the problem of determining what to learn, and whether particular GVFs are useful for informing decision-making.

There is no metric the automate the analysis of prediction usefulness for decision-making. Consequently, system designers rely on the metrics currently avialable: namely, prediction error; designers are missing metrics that summarise and evaluate the usefulness of features. In the following section, we propose a metric to fill this gap: evaluation of prediction usefulness by examining a value function's internal learning process.


\section{Proposal: evaluate feature relevance}

We demonstrated that error in isolation of any additional information is misleading: empirical return error and RUPEE are insufficient to determine whether a model is useful for informing down-stream decision-making by an agent. This inability to assess the usefulness of predictions is a major hurdle, the purpose of constructing knowledge is it's use in supporting decision making. If measures of accuracy verified using data available to the agent are not enough to asses the usefulness of a model, what should a designer do?

We need not only look at signals external to the agent for clues about performance: we can also look inwards and examine the learning process to assess an agent's knowledge---how the agent is modifying its parameters. Examining an agent's parameters is not unusual. For example, Unexpected Demon Error (UDE), can used to gauge how `surprising' a given observation is to an agent~\cite{white_developing_2015}. By examining the surprise, we can gauge how current experience relates to past experiences---e.g., detecting faults in a system~\cite{gunther_predictions_2018}.

Similarly, there are many such parameters that an agent can modify during learning, and that modification can be monitored. Of particular interest are meta-learning methods: higher-order learning processes that modify the learning parameters of an agent (e.g., IDBD~\cite{sutton_adapting_1992}). One notable example is step-size (learning rate) adaptation.

Some meta-learning methods tune the step-size parameter $\alpha$ based on the relevance of a given feature. For instance, TD Incremental Delta-Bar-Delta (TIDBD)~\cite{kearney_learning_2019} assigns a step-size $\alpha_i$ to each weight $w_i$, adjusting the step-size based on the correlation of recent weight updates. If many weight updates are made in the same direction, then a more efficient use of data would have been to make one large update with a larger $\alpha_i$. If an update has over-shot, then the weight updates will be uncorrelated, and thus the step-size should be smaller. More broadly, we can view these forms of step-size adaptation as the most basic form of representation learning.

All else being equal, a good model is one whose features are well aligned with the prediction problem at hand: the features are relevant and are aligned with the prediction task. Even in early-learning where an agent is adjusting its model, or in situations where non-stationarity in the environment may introduce unexpected error, if the features are relevant to the prediction task we can expect a reasonable model in the limit. One way to determine the relevance of features is by learning step-sizes.

\subsection{Derivation of off-policy TIDBD}

To demonstrate how step-sizes as feature relevance can be informative, we generalize TIDBD~\cite{kearney_learning_2019} to GTD ($\lambda$), creating a step-size adaptation method suited for the off-policy \texttt{touch}, \texttt{touch-left}, and \texttt{touch-right} predictions we previously introduced. Off-policy AutoStep for GTD adds a few additional memory parameters to perform step-size adaptation.

Here, we derive the relevant updates as follows. TIDBD minimises $\delta^{2}$ the squared TD error with respect to meta-weights $\beta$ that specify the agent's step-size on each time-step.

\begin{equation}
\begin{split}
    \beta_{i,t+1} &= \beta_{i,t} - \frac{1}{2} \theta \frac{\partial \delta_{t}^2}{\partial \beta_i} \\
    &= \beta_{i,t} - \frac{1}{2} \theta \sum_j \frac{\partial \delta_{t}^2}{\partial w_{j,t}}\frac{\partial w_j}{\partial \beta_i}\\
\end{split}
\end{equation}

We expand $\frac{\partial \delta_{t}^2}{\partial \beta_i}$ using the chain-rule. 
As in~\cite{sutton_adapting_1992}, we make the assumption that the effect of changing the step size $\alpha_i = \exp(\beta_i)$ for some feature $\phi_{i,t}$ will predominantly be on the weight $w_i$.

\begin{equation}
   \beta_{i,t+1} \approx \beta_{i,t} -\frac{1}{2}\theta \frac{\partial \delta_{t}^2}{\partial w_{i,t}} \frac{\partial w_{i,t}}{\partial \beta_i}
\end{equation}

We are minimizing the TD error $\delta = c_{t+1} + \gamma V(\phi_{t+1} - V(\phi))$, where $c$ is the cumulant, $\gamma$ is the discount factor, and $V$ is our value function, and $\phi$ is the state as constructed by a function approximator. Given $\delta$ is a biased estimate of the error, dependent on our value function $V$, we take the semi-gradient $-V(\phi_{t})$.

\begin{equation}
    \begin{split}
       -\frac{1}{2} \frac{\partial \delta_{t}^2}{\partial w_{i,t}} &=  - \delta \frac{\partial [-V(\phi_{i,t})]}{\partial w_{i,t}}\\ 
       &= \delta_{t}\phi_{i,t}
    \end{split}
\end{equation}

\begin{equation}
    \beta_{i,t+1} \approx \beta_{i,t} + \delta_{t} \phi_{i,t} \frac{\partial w_{i,t}}{\partial \beta_i}
\end{equation}

We then describe $\frac{\partial w_{i,t}}{\partial \beta_i}$ as $\omega$. GTD($\lambda$) updates the weights as $w\gets w+\alpha[\delta e - \gamma (1-\lambda)(e^\top h) \phi_{t+1}]$. We can then write the update to $\omega$ recursively as follows:

\begin{equation}
  \begin{split}
        &\omega_{t+1} = \frac{\partial}{\partial \beta}\Big[w + \alpha(\delta e - \gamma(1-\lambda)\phi_{t+1}e_{t}^\top h_{t})\Big]\\
    &= \omega_{t} + \alpha \delta e + \alpha e \frac{\partial}{\partial \beta}[\delta] + \alpha \delta \frac{\partial}{\partial \beta}[e] \\
    &- \alpha \gamma (1 - \lambda)\phi_{t+1} e^\top h - \alpha \gamma (1-\lambda)\phi_{t+1} \frac{\partial}{\partial \beta}[e^\top h]\\
    &\approx \omega_{t} + \alpha \delta e - \alpha \omega_{t} \phi_{t} e - \alpha \gamma (1 - \lambda) \phi_{t+1} e^\top h \\
    &- \alpha \gamma (1-\lambda)\phi_{t+1} e^\top \frac{\partial}{\partial \beta}[h]\\
    &= \omega_{t} + \alpha \bigg( \delta e - \omega_{t}\phi_{t}e - \gamma(1-\lambda) \phi_{t+1}(e^\top h + e^\top \eta_{t}) \bigg)
    \end{split}
\end{equation}

In GTD ($\lambda$) the bias-correction updated update is $h \gets h + \alpha (\delta e - (h^\top \phi_{t})\phi_{t})$. Similar to $\omega$, we define $\frac{\partial h_{t}}{\partial \beta}$ as $\eta$. The $\eta$ update is as follows:

\begin{equation}
    \begin{split}
        &\eta_{t+1} = \frac{\partial}{\partial \beta}\Big[h_{t} + \alpha (\delta e - (h^\top \phi_{t}) \phi_{t})\Big]\\
        &= \eta_{t} + \alpha \delta e + \alpha \frac{\partial}{\partial \beta} [\delta] e + \alpha \delta \frac{\partial}{\partial \beta}[e] - \alpha (h^\top \phi_{t}) \phi_{t} \\
        & - \alpha \frac{\partial}{\partial \beta}(h^\top \phi_{t}) \phi_{t} \\
        &\approx \eta_{t} + \alpha \delta e - \alpha \omega_{t}\phi_{t} e -\alpha (h^\top \phi_{t}) \phi_{t} - \alpha (\eta_{t}^\top \phi_{t}) \phi_{t} \\    
    \end{split}
\end{equation}
    
We now have our three additional updates defined for GTD ($\lambda$) IDBD. This results in our GTD IDBD. We now have all the features for a GTD version of IDBD.

\begin{equation}
    \beta \gets \beta + \theta \delta \phi_{t} \omega_{t}
\end{equation}

\begin{equation}
    \eta \gets \eta+ \alpha \bigg( \Big(e\big( \delta - \omega_{t}\big)- \big( h + \eta \big)^\top \phi_{t}\Big) \phi_{t} \bigg)
\end{equation} 

\begin{equation}
    \begin{split}
             \omega \gets \omega + \alpha \bigg(e\big( \delta - \omega_{t}\phi_{t}\big)   - \gamma\phi_{t+1}(1-\lambda)e^\top(h +\eta) \bigg)
    \end{split}
\end{equation}

To generalize AutoStep~\cite{mahmood_tuning-free_2012} to GTD($\lambda$) we need two more additions to GTD($\lambda$): 1) a running average of meta-weight updates to prevent instability in our meta-weights caused by dramatic changes in the target of the underlying learning method, and 2) a normalization by the \textit{effective step size} to prevent over-shooting on an individual example.

The effective step size describes the amount by which the error has been reduced on a particular example after a weight update. If the effective step-size is greater than one, then we have over-shot on a particular example. To prevent over-shooting, we divide the step-size on each time-step by $\max(1,  \frac{\delta_t(t) - \delta_{t+1}(t)}{\delta_t(t)})$ To find the effective step-size, we simplify the following:

\begin{equation}
    \begin{split}
        &\frac{\delta_t(t) - \delta_{t+1}(t)}{\delta_t(t)} = \frac{1}{\delta_t(t)}-\Big[\Big(C_{t+1} + \gamma V_t(\phi_{t+1}) - V_{t}(\phi_t) \Big) \\
        &\phantom{\frac{\delta_t(t) - \delta_{t+1}(t)}{\delta_t(t)} = }\Big(C_{t+1} + \gamma V_{t+1}(\phi_{t+1}) - V_{t+1}(\phi_t)\Big)\Big] \\
        &= \frac{1}{\delta_t(t)}\Big[\Big(\gamma V_t(\phi_{t+1}) - V_{t}(\phi(t)) \Big) -\\
        &\phantom{\frac{1}{\delta_t(t)}\Big[\Big(\gamma V_{t}} \Big(\gamma V_{t+1}(\phi_{t+1}) - V_{t+1}(\phi_{t})\Big)\Big] \\
    \end{split}
\end{equation}
    
We simplify to the resulting effective step size:

\begin{equation}
   \begin{split}
 \Big[\alpha e - \frac{\gamma (1-\lambda) \phi_{t+1}e^\top h }{\delta}\Big]^\top \Big[\phi_{t} &- \gamma \phi_{t+1}\Big]    
   \end{split}
\end{equation}
    
Having found the effective step-size, we must define an update normalizer. On each time-step IDBD updates the step-sizes by $\delta \phi \omega$. We take a decaying trace of the maximum weight update, $\xi \gets \max(|\delta\phi\omega|, \xi + \frac{1}{\tau}\alpha \phi e (|\delta \phi \omega| - \xi))$, where $\tau$ is a parameter that specifies how quickly $\xi$ decays. This has the effect of maintaining a decaying trace of the maximum update such that a large change in the underlying learning target does not lead to instability in the step-size parameter update.

\begin{algorithm}[b]
\caption{GTD ($\lambda$) with AutoStep step-size tuning.}\label{IDBD}
\begin{algorithmic}[1]
\STATE Initialize vectors $\omega$, $\eta$, $\alpha$, $e$, $\xi$, and $w$ of size $n$ number of features. Set $\tau$ as a decay value e.g., $10^{4}$ and $\theta$ as a meta step-size (e.g., $10^{-2}$). Observe initial state $\phi$, take initial action $a$ and observe next state $\phi^\prime$ and cumulant $c$.\\
\STATE Repeat for each following $\phi^\prime$, $a$ pair:

  \INDSTATE[1] $\delta \gets c + \gamma w^\top\phi^\prime - w^\top\phi_t$
  		\INDSTATE[1] $\xi \gets \max($
    		\INDSTATE[2] $|\delta \phi \omega|, $
     		\INDSTATE[2] $\xi + \frac{1}{\tau} \alpha \phi e (| \delta \phi \omega | - \xi )$
    	\INDSTATE[1] $)$
   \INDSTATE[1] For $i = 1, 2, \ldots, n $:
        \INDSTATE[2] if $\xi_i \neq 0$:
    		\INDSTATE[3] $\alpha \gets \alpha \exp(\theta \frac{\delta \phi \omega}{\xi_i} )$
  	\INDSTATE [1] $M \gets \max( $
    \INDSTATE[2] $\Big[\alpha e - \frac{\gamma (1-\lambda) \phi_{t+1}e^\top h }{\delta}\Big]^\top \Big[\phi_{t} - \gamma \phi_{t+1}\Big], 1$
    \INDSTATE[1] $)$
   	\INDSTATE[1] $\alpha \gets \frac{\alpha}{M}$
    \INDSTATE[1] $\rho \gets \frac{\pi(\phi, a)}{\mu(\phi, a)}$
  	\INDSTATE[1] $w \gets w+ \alpha(\delta e - \gamma (1-\lambda) e ^\top h \phi^{\prime})$
  	\INDSTATE[1] $h \gets h + \alpha (\delta e - (h^\top \phi)\phi)$
  	\INDSTATE[1] $e \gets \rho(e \gamma \lambda + \phi)$
    \INDSTATE[1] $\omega \gets \omega + \alpha \bigg(e\big( \delta - \omega \phi\big)   - \gamma\phi^{\prime} (1-\lambda)e^\top(h +\eta) \bigg)$
  	\INDSTATE[1] $\eta \gets \eta+ \alpha \bigg( \Big(e\big( \delta - \omega_{t}\big)- \big( h + \eta \big)^{\top}\phi \Big) \phi \bigg)$
    \INDSTATE[1] $\phi \gets \phi^\prime$
\end{algorithmic}\label{TIDBD_alg}
\end{algorithm}

These updates can then be combined with the underlying GTD($\lambda$) updates to produce and Autostep GTD ($\lambda$) (Algorithm~\ref{TIDBD_alg})

\subsection{Evaluating via feature relevance}

\begin{figure*}[h]
\centering
    \begin{subfigure}[b]{0.45\linewidth}
    \centering
    \includegraphics[width=\linewidth, height=2.5in, keepaspectratio]{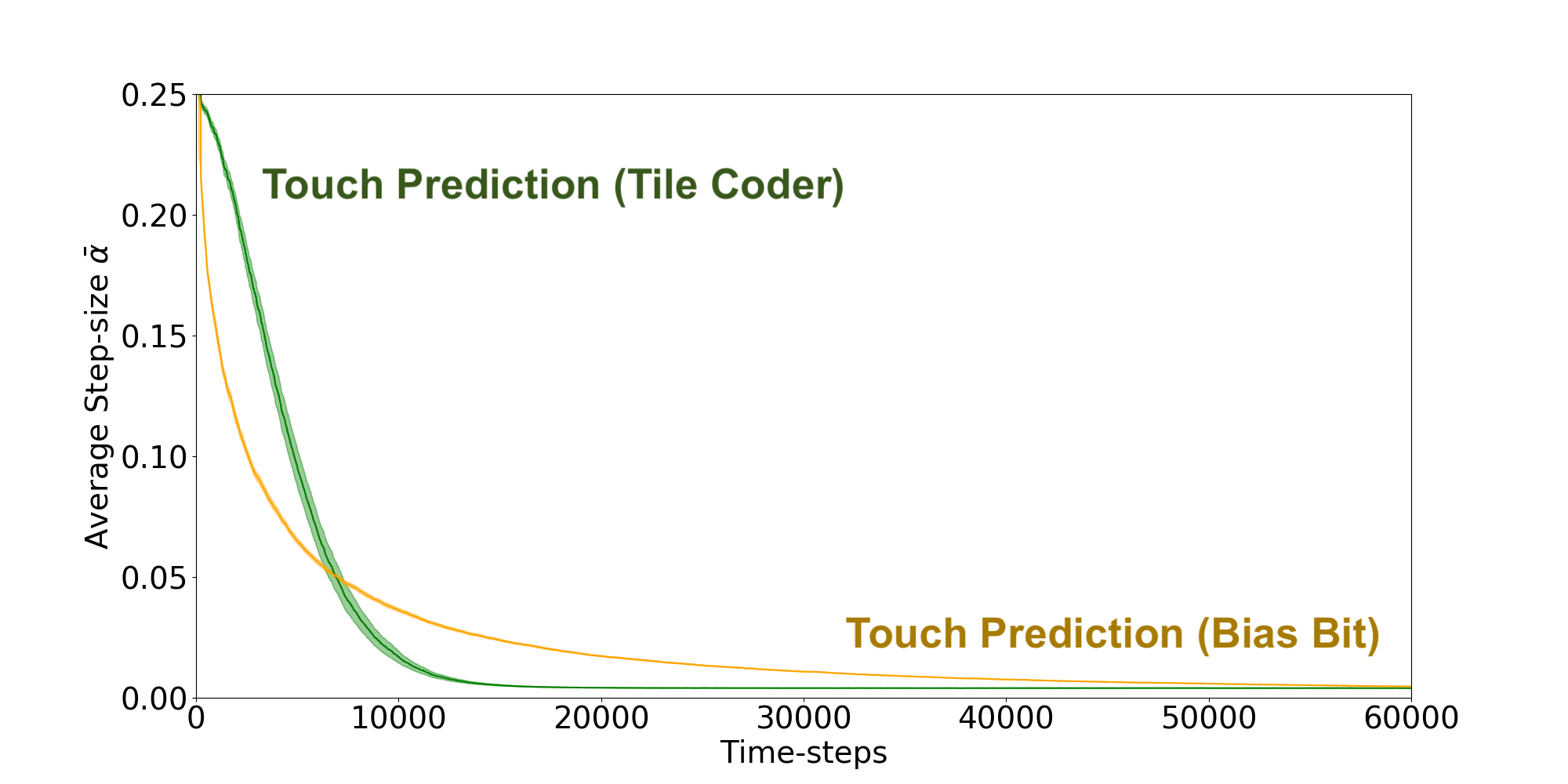}
    \caption{The average active step-size for both \texttt{touch} predictions. Anticipatory prediction in green; tracking based prediction in orange. \\}\label{step-size-l1}
    \end{subfigure}%
    \hspace{1em}%
    \begin{subfigure}[b]{0.45\linewidth}
    \centering
    \includegraphics[width=\linewidth, height=2.5in, keepaspectratio]{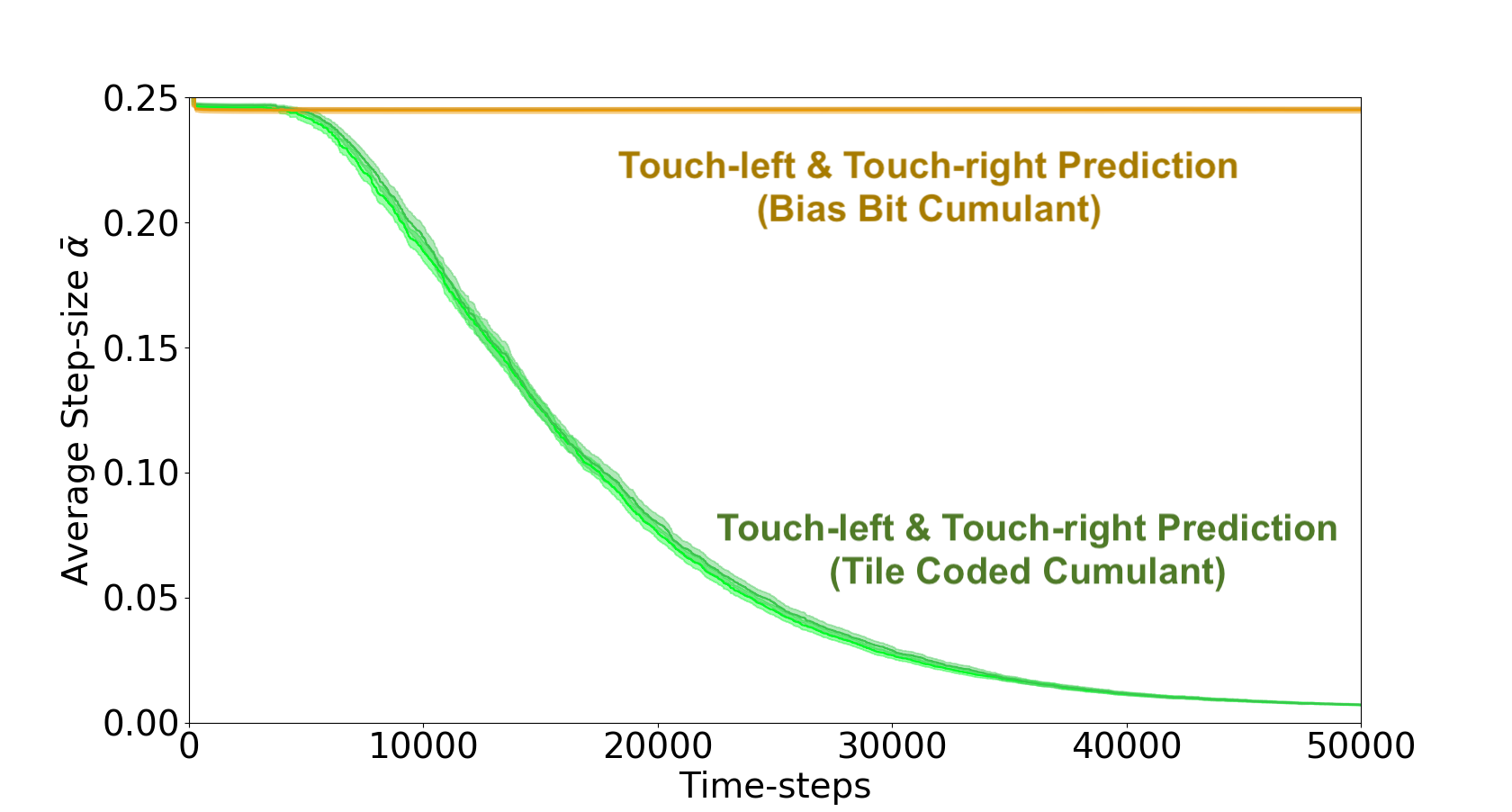}
    \caption{Average active step-size for the \texttt{touch-left} and \texttt{touch-right} predictions. Anticipatory predictions in green; tracking-based predictions in orange. }\label{step-size-l2}
    \end{subfigure}
    \caption{The average active step-sizes for each layer of both the prediction and tracking networks averaged over 30 independent trials. Error bars are standard error of the mean.}\label{step-sizes}

    \begin{subfigure}[b]{0.45\linewidth}
    \centering
    \includegraphics[width=\linewidth, height=2.5in, keepaspectratio]{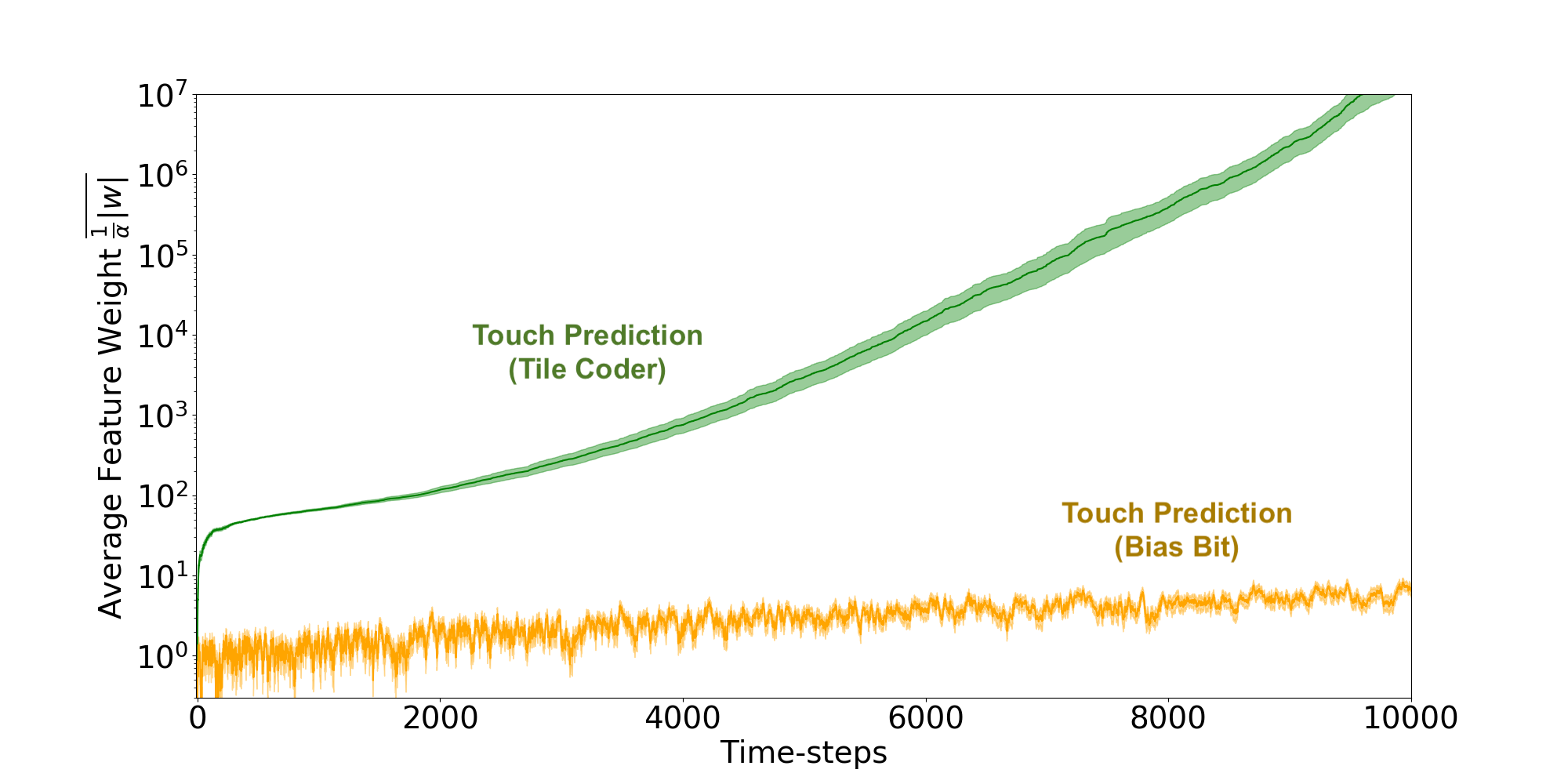}
    \caption{Average weighted feature relevance $\overline{\frac{1}{\alpha }|w|}$ for \texttt{touch} predictions. Anticipatory tile-coded prediction in green; tracking bias-bit prediction in orange.}\label{step-size-weigths-l1}
    \end{subfigure}%
    \hspace{1em}%
    \begin{subfigure}[b]{0.45\linewidth}
    \centering
    \includegraphics[width=\linewidth, height=2.5in, keepaspectratio]{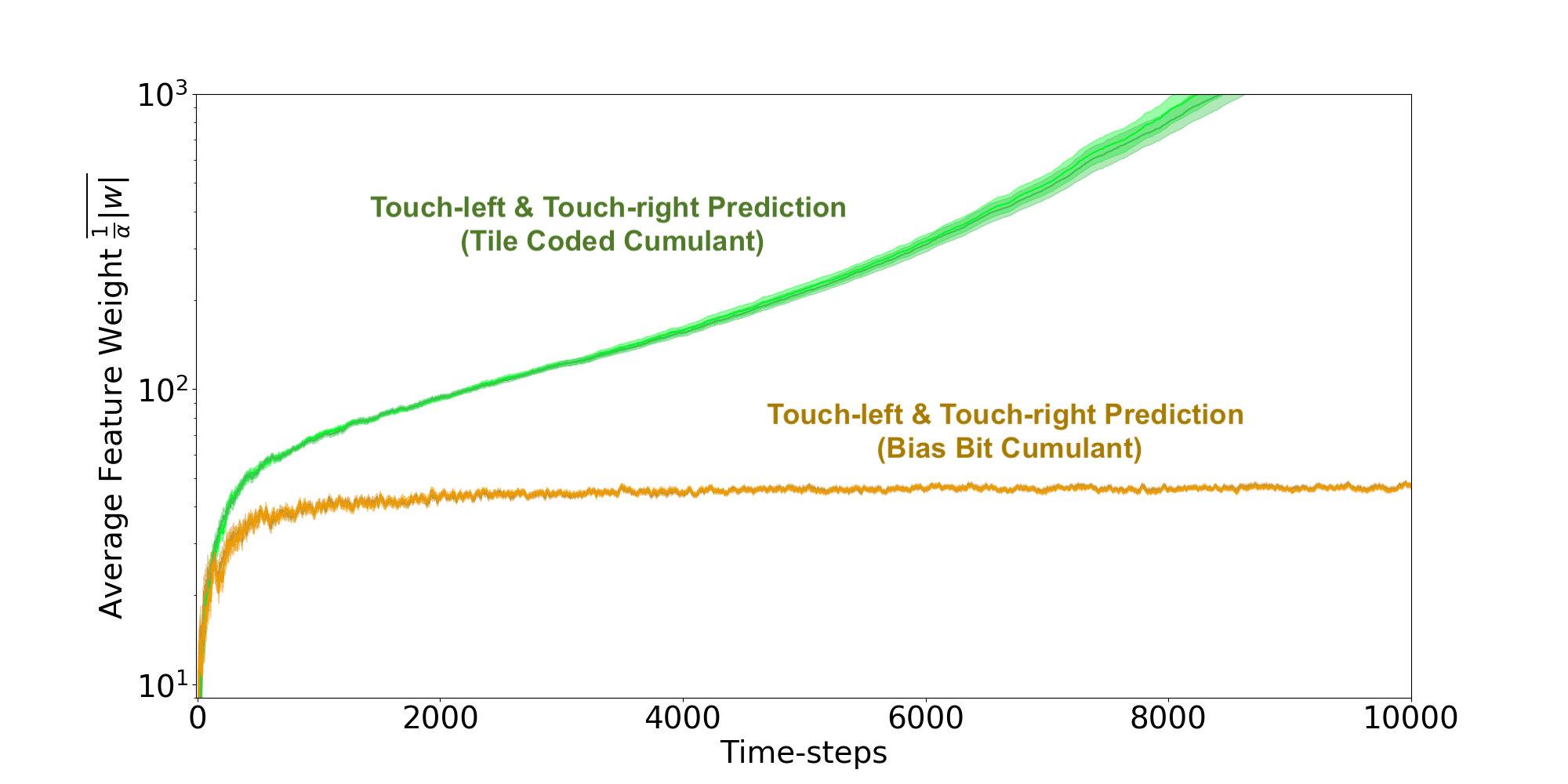}
    \caption{Average weighted feature relevance $\overline{\frac{1}{\alpha }|w|}$ for the touch-left and touch-right predictions. Anticipatory predictions in green; tracking-based predictions in orange. }\label{step-size-weights-l2}

    \end{subfigure}
    \caption{The average weighted feature relevance $\overline{\frac{1}{\alpha }|w|}$ for each layer of both the prediction and tracking networks. Each is run over 30 independent trials. Error bars are standard error of the mean.}\label{step-sizes-weights}
\end{figure*}

Having generalised TIDBD to GTD ($\lambda$), we now return to the MineCraft domain and perform the same experiments, now using step-size adaptation. In figure~\ref{step-sizes} the average active\footnote{For our function approximator, we use a tile-coder. The tile-coder outputs a binary feature vector---only a portion of all features are active on a given time-step. We multiple the average absolute step-size by the number of active features so that two function approximators with differing active feature sizes will have equivalent scale and can be compared.} step-size value for the duration of the experiment is depicted. As was the case in the prior experiments, we have two agents each learning three predictions: \texttt{touch}, \texttt{touch-right}, and \texttt{touch-left}. One agent has a representation sufficient to learn the underlying \texttt{touch} prediction with reasonable accuracy (green), while the other does not (orange).

By examining the step-size values, we are able to discriminate between the tracking and predictive \texttt{touch-left} and \texttt{touch-right} predictions (Figure~\ref{step-size-l2}); however, we find that the tracking and predictive \texttt{touch} predictions are not appreciably different when examining their step sizes late in learning progress (as shown in Figure~\ref{step-size-l1}). Independent of learned weights, step-sizes do not tell the full story; our step-sizes $\alpha$ are a weighting of our features $\phi$ when learning some weights $w$. The learned step-sizes $\alpha$ in combination with the learned weights $w$ give us greater insight into the performance of our GVFs. In Figure~\ref{step-sizes-weights} a combination of the absolute value of the learned weights and step-sizes are plotted: $\overline{\frac{1}{\alpha}|w|}$. We take $\frac{1}{\alpha}$, as the magnitude of the step-size describes progress in learning. Intuitively, a feature which is stable, and thus has a small $\alpha_i$, and has a relatively large weight $w_i$ is preferable.

By examining the learned step-sizes and weights $\overline{\frac{1}{\alpha}|w|}$, we are finally able to separate the tracking and anticipatory \texttt{touch} predictions using an easily calculated metric (Figure~\ref{step-size-weights-l2}). As the step-sizes decrease, the value of both the tracking and anticipatory predictions rises; however, since the magnitude of the weight $w$ is low for the bias-bit, its weighted feature value remains low. This clarity in comparison carries over to the \texttt{touch-left} and \texttt{touch-right} predictions (Figure~\ref{step-size-weights-l2}). From Figure~\ref{step-size-l2}, we know that the tracking-based \texttt{touch-left} and \texttt{touch-right} predictions' step-sizes never decay---the tracking predictions' step-sizes maintain an average value of approximately 0.25 for the duration of the trials, while the anticipatory predictions' step-sizes decay as the predictions are learnt. This results in a pronounced bifurcation between the two predictions. By looking at weighted features, we are able see and interpret what has been lost in our error estimates.

The practice of using step-sizes that describe feature relevance to inform other aspects of learning is already an established practice. For instance, learned step-sizes have been used to inform feature discovery~\cite{mahmood_representation_2013}, and exploration methods~\cite{linke2020adapting}. Recent work has suggested that step-sizes can be used to monitor the status of robots and indicate when physical damage has occurred to a system~\cite{gunther_meta-learning_2019,gunther_examining_2020}. Moreover, using internal learning measurements to evaluate predictive knowledge systems has been suggested in other works~\cite{sherstan_introspective_2016}, although no existing applications of predictive knowledge use step-sizes for evaluation.

Using the learning method we generalized, AutoStep for GTD ($\lambda$), we can learn step-sizes online and incrementally as the agent is interacting with the environment. In situations where traditional prediction error metrics fail, the magnitude of learned weights and step-sizes enables differentiation  between GVFs that are useful in informing further predictions, and GVFs which are not. {\bf In brief, we show that GVFs can be evaluated in a meaningful, scalable way using feature relevance.}

\section{Conclusion}

Agents often benefit from constructing general knowledge of their world. How the models that compose this knowledge are constructed and evaluated is a challenging open problem. In this paper, we critically discussed a common way of evaluating an agent's knowledge: model accuracy with respect to observed values. As a first primary contribution of this work, we demonstrated how strict measures of accuracy can be misleading. We further showed how critical areas of performance can be hidden by biased measures of error, leading to a poor choice of model. Building on this observation, we next demonstrated how poor evaluation in learned models can lead to more serious errors in down-stream learning tasks (e.g, prediction) which depend on these models. As a final contribution, we proposed an alternative evaluation approach that instead examines an agent's learned parameters as a basis for certifying learned knowledge, specifically focusing on learned weights and step-size values. Using these additional sources of information, we showed that we are able to differentiate between useful and useless model in a setting that was indistinguishable when using standard error or accuracy-based assessments. This paper therefore contributes a first look into how predictive models evaluation and use are related. Decoupling the evaluation of predictions from strict measures of accuracy is an key step towards building general, modular representations of knowledge.

\begin{acks}
The authors thank Joseph Modayil, Jaden Travnik, Johannes Gunther, Brian Tanner, and Katya Kudashkina for detailed reading of this manuscript and a number of valuable suggestions and discussions. This research was undertaken, in part, thanks to funding from the Canada Research Chairs program, the Canada CIFAR AI Chairs program, the Canada Foundation for Innovation, the Alberta Machine Intelligence Institute, Alberta Innovates, and the Natural Sciences and Engineering Research Council. AKK was supported by scholarships and awards from NSERC, Alberta Innovates, and Borealis AI.\end{acks}


\bibliographystyle{SageH}
\bibliography{bibliography}

\begin{thebibliography}{41}
\providecommand{\natexlab}[1]{#1}
\providecommand{\url}[1]{\texttt{#1}}
\providecommand{\urlprefix}{URL }
\expandafter\ifx\csname urlstyle\endcsname\relax
  \providecommand{\doi}[1]{DOI:\discretionary{}{}{}#1}\else
  \providecommand{\doi}{DOI:\discretionary{}{}{}\begingroup
  \urlstyle{rm}\Url}\fi

\bibitem[{Barreto et~al.(2017)Barreto, Dabney, Munos, Hunt, Schaul, {van
  Hasselt} and Silver}]{barreto_successor_2017}
Barreto A, Dabney W, Munos R, Hunt JJ, Schaul T, {van Hasselt} HP and Silver D
  (2017) Successor features for transfer in reinforcement learning.
\newblock In: \emph{Advances in Neural Information Processing Systems}. pp.
  4055--4065.

\bibitem[{Clark(2013)}]{clark_whatever_2013}
Clark A (2013) Whatever next? {{Predictive}} brains, situated agents, and the
  future of cognitive science.
\newblock \emph{Behavioral and brain sciences} 36(3): 181--204.

\bibitem[{Comanici et~al.(2018)Comanici, Precup, Barreto, Toyama, Ayg\&\#xFC,
  {n}, Hamel, Vezhnevets, Hou and Mourad}]{comanici_knowledge_2018}
Comanici G, Precup D, Barreto A, Toyama DK, Ayg\&\#xFC E, {n}, Hamel P,
  Vezhnevets S, Hou S and Mourad S (2018) Knowledge {{Representation}} for
  {{Reinforcement Learning}} using {{General Value Functions}}.
\newblock \emph{technical report} .

\bibitem[{Edwards et~al.(2016)Edwards, Hebert and
  Pilarski}]{edwards_machine_2016}
Edwards AL, Hebert JS and Pilarski PM (2016) Machine learning and unlearning to
  autonomously switch between the functions of a myoelectric arm.
\newblock In: \emph{Biomedical {{Robotics}} and {{Biomechatronics}}
  ({{BioRob}}), 2016 6th {{IEEE International Conference}} On}. {IEEE}, pp.
  514--521.

\bibitem[{Gilbert(2009)}]{gilbert_stumbling_2009}
Gilbert D (2009) \emph{Stumbling on Happiness}.
\newblock {Vintage Canada}.

\bibitem[{G{\"u}nther et~al.(2020)G{\"u}nther, Ady, Kearney, Dawson and
  Pilarski}]{gunther_examining_2020}
G{\"u}nther J, Ady NM, Kearney A, Dawson MR and Pilarski PM (2020) Examining
  the {{Use}} of {{Temporal}}-{{Difference Incremental
  Delta}}-{{Bar}}-{{Delta}} for {{Real}}-{{World Predictive Knowledge
  Architectures}}.
\newblock \emph{Frontiers in Robotics and AI} 7: 34.

\bibitem[{G{\"u}nther et~al.(2019)G{\"u}nther, Kearney, Ady, Dawson and
  Pilarski}]{gunther_meta-learning_2019}
G{\"u}nther J, Kearney A, Ady N, Dawson MR and Pilarski PM (2019) Meta-learning
  for {{Predictive Knowledge Architectures}}: {{A Case Study Using TIDBD}} on a
  {{Sensor}}-rich {{Robotic Arm}}.
\newblock In: \emph{Proc. of the 18th {{International Conference}} on
  {{Autonomous Agents}} and {{Multiagent Systems}} ({{AAMAS}} 2019)}.
  {Montreal, Canada}, pp. 1967--1969.

\bibitem[{G{\"u}nther et~al.(2018)G{\"u}nther, Kearney, Dawson, Sherstan and
  Pilarski}]{gunther_predictions_2018}
G{\"u}nther J, Kearney A, Dawson MR, Sherstan C and Pilarski PM (2018)
  Predictions, surprise, and predictions of surprise in general value function
  architectures.
\newblock In: \emph{{{AAAI}} 2018 {{Fall Symposium}} on {{Reasoning}} and
  {{Learning}} in {{Real}}-{{World Systems}} for {{Long}}-{{Term Autonomy}}}.
  pp. 22--29.

\bibitem[{G{\"u}nther et~al.(2016)G{\"u}nther, Pilarski, Helfrich, Shen and
  Diepold}]{gunther_intelligent_2016}
G{\"u}nther J, Pilarski PM, Helfrich G, Shen H and Diepold K (2016) Intelligent
  laser welding through representation, prediction, and control learning:
  {{An}} architecture with deep neural networks and reinforcement learning.
\newblock \emph{Mechatronics} 34: 1--11.

\bibitem[{Ha and Schmidhuber(2018)}]{ha_world_2018}
Ha D and Schmidhuber J (2018) Recurrent world models facilitate policy
  evolution.
\newblock In: \emph{Advances in Neural Information Processing Systems},
  volume~31. pp. 2450--2462.

\bibitem[{Jaderberg et~al.(2017)Jaderberg, Mnih, Czarnecki, Schaul, Leibo,
  Silver and Kavukcuoglu}]{jaderberg_reinforcement_2016}
Jaderberg M, Mnih V, Czarnecki WM, Schaul T, Leibo JZ, Silver D and Kavukcuoglu
  K (2017) Reinforcement learning with unsupervised auxiliary tasks.
\newblock In: \emph{5th International Conference on Learning Representations,
  {ICLR} 2017, Toulon, France, April 24-26, 2017, Conference Track
  Proceedings}.

\bibitem[{Johnson et~al.(2016)Johnson, Hofmann, Hutton and
  Bignell}]{johnson2016malmo}
Johnson M, Hofmann K, Hutton T and Bignell D (2016) The malmo platform for
  artificial intelligence experimentation.
\newblock In: \emph{IJCAI}. Citeseer, pp. 4246--4247.

\bibitem[{Kearney et~al.(2019)Kearney, Veeriah, Travnik, Pilarski and
  Sutton}]{kearney_learning_2019}
Kearney A, Veeriah V, Travnik J, Pilarski PM and Sutton RS (2019) Learning
  {{Feature Relevance Through Step Size Adaptation}} in
  {{Temporal}}-{{Difference Learning}}.
\newblock \emph{arXiv preprint arXiv:1903.03252} .

\bibitem[{Koop(2008)}]{koop_investigating_2008}
Koop A (2008) \emph{Investigating Experience: {{Temporal}} Coherence and
  Empirical Knowledge Representation}.
\newblock {{Msc Thesis}}, University of Alberta.

\bibitem[{Linke et~al.(2020)Linke, Ady, White, Degris and
  White}]{linke2020adapting}
Linke C, Ady NM, White M, Degris T and White A (2020) Adapting behavior via
  intrinsic reward: A survey and empirical study.
\newblock \emph{Journal of Artificial Intelligence Research} 69: 1287--1332.

\bibitem[{Mahmood and Sutton(2013)}]{mahmood_representation_2013}
Mahmood AR and Sutton RS (2013) Representation {{Search}} through {{Generate}}
  and {{Test}}.
\newblock In: \emph{{{AAAI Workshop}}: {{Learning Rich Representations}} from
  {{Low}}-{{Level Sensors}}}.

\bibitem[{Mahmood et~al.(2012)Mahmood, Sutton, Degris and
  Pilarski}]{mahmood_tuning-free_2012}
Mahmood AR, Sutton RS, Degris T and Pilarski PM (2012) Tuning-free step-size
  adaptation.
\newblock In: \emph{Acoustics, {{Speech}} and {{Signal Processing}}
  ({{ICASSP}}), 2012 {{IEEE International Conference}} On}. {IEEE}, pp.
  2121--2124.

\bibitem[{Modayil and Sutton(2014)}]{modayil_prediction_2014}
Modayil J and Sutton RS (2014) Prediction driven behavior: {{Learning}}
  predictions that drive fixed responses.
\newblock In: \emph{The {{AAAI}}-14 {{Workshop}} on {{Artificial Intelligence}}
  and {{Robotics}}, {{Quebec City}}, {{Quebec}}, {{Canada}}}.

\bibitem[{Modayil et~al.(2014)Modayil, White and
  Sutton}]{modayil_multi-timescale_2014}
Modayil J, White A and Sutton RS (2014) Multi-timescale nexting in a
  reinforcement learning robot.
\newblock \emph{Adaptive Behavior} 22(2): 146--160.

\bibitem[{N{\"o}e(2004)}]{noe_action_2004}
N{\"o}e A (2004) \emph{Action in Perception}.
\newblock {MIT press}.

\bibitem[{Pezzulo(2011)}]{pezzulo_grounding_2011}
Pezzulo G (2011) Grounding procedural and declarative knowledge in sensorimotor
  anticipation.
\newblock \emph{Mind \& Language} 26(1): 78--114.

\bibitem[{Pezzulo et~al.(2013)Pezzulo, Donnarumma and
  Dindo}]{pezzulo_human_2013}
Pezzulo G, Donnarumma F and Dindo H (2013) Human sensorimotor communication:
  {{A}} theory of signaling in online social interactions.
\newblock \emph{PloS one} 8(11): e79876.

\bibitem[{Pilarski et~al.(2012)Pilarski, Dawson, Degris, Carey and
  Sutton}]{pilarski_dynamic_2012}
Pilarski PM, Dawson MR, Degris T, Carey JP and Sutton RS (2012) Dynamic
  switching and real-time machine learning for improved human control of
  assistive biomedical robots.
\newblock In: \emph{Biomedical {{Robotics}} and {{Biomechatronics}}
  ({{BioRob}}), 2012 4th {{IEEE RAS}} \& {{EMBS International Conference}} On}.
  {IEEE}, pp. 296--302.

\bibitem[{Pilarski and Sherstan(2016)}]{pilarski_steps_2016}
Pilarski PM and Sherstan C (2016) Steps toward knowledgeable neuroprostheses.
\newblock In: \emph{Biomedical {{Robotics}} and {{Biomechatronics}}
  ({{BioRob}}), 2016 6th {{IEEE International Conference}} On}. {IEEE}, pp.
  220--220.

\bibitem[{Rao and Ballard(1999)}]{rao_predictive_1999}
Rao RP and Ballard DH (1999) Predictive coding in the visual cortex: A
  functional interpretation of some extra-classical receptive-field effects.
\newblock \emph{Nature neuroscience} 2(1): 79--87.

\bibitem[{Ring(2016)}]{ring_representing_2016}
Ring M (2016) Representing {{Knowledge}} as {{Predictions}} (and {{State}} as
  {{Knowledge}}).

\bibitem[{Ring(1997)}]{ring_child_1997}
Ring MB (1997) {{CHILD}}: {{A}} first step towards continual learning.
\newblock \emph{Machine Learning} 28(1): 77--104.

\bibitem[{Schlegel et~al.(2018)Schlegel, White, Patterson and
  White}]{schlegel_general_2018}
Schlegel M, White A, Patterson A and White M (2018) General value function
  networks.
\newblock \emph{arXiv:1807.06763 [cs, stat]} .

\bibitem[{Sherstan et~al.(2020)Sherstan, Dohare, MacGlashan, G{\"u}nther and
  Pilarski}]{sherstan_work_2020}
Sherstan C, Dohare S, MacGlashan J, G{\"u}nther J and Pilarski PM (2020)
  Gamma-nets: Generalizing value estimation over timescale.
\newblock In: \emph{Proceedings of the AAAI Conference on Artificial
  Intelligence}, volume~34. pp. 5717--5725.

\bibitem[{Sherstan et~al.(2018)Sherstan, Machado and
  Pilarski}]{sherstan_accelerating_2018}
Sherstan C, Machado MC and Pilarski PM (2018) Accelerating {{Learning}} in
  {{Constructive Predictive Frameworks}} with the {{Successor Representation}}.
\newblock \emph{arXiv preprint arXiv:1803.09001} .

\bibitem[{Sherstan et~al.(2016)Sherstan, White, Machado and
  Pilarski}]{sherstan_introspective_2016}
Sherstan C, White A, Machado MC and Pilarski PM (2016) Introspective agents:
  {{Confidence}} measures for general value functions.
\newblock In: \emph{International {{Conference}} on {{Artificial General
  Intelligence}}}. {Springer}, pp. 258--261.

\bibitem[{Sherstov and Stone(2005)}]{sherstov2005function}
Sherstov AA and Stone P (2005) Function approximation via tile coding:
  Automating parameter choice.
\newblock In: \emph{International Symposium on Abstraction, Reformulation, and
  Approximation}. Springer, pp. 194--205.

\bibitem[{Singh et~al.(2005)Singh, Barto and
  Chentanez}]{singh2005intrinsically}
Singh S, Barto AG and Chentanez N (2005) Intrinsically motivated reinforcement
  learning.
\newblock Technical report, Massachusetts University Amherst Dept Of Computer
  Science.

\bibitem[{Sutton(1988)}]{sutton_learning_1988}
Sutton RS (1988) Learning to predict by the methods of temporal differences.
\newblock \emph{Machine Learning} 3(1): 9--44.
\newblock \doi{10.1007/BF00115009}.

\bibitem[{Sutton(1992)}]{sutton_adapting_1992}
Sutton RS (1992) Adapting bias by gradient descent: {{An}} incremental version
  of delta-bar-delta.
\newblock In: \emph{AAAI}. pp. 171--176.

\bibitem[{Sutton and Barto(2019)}]{sutton_reinforcement_1998}
Sutton RS and Barto AG (2019) \emph{Reinforcement Learning: {{An}}
  Introduction}.
\newblock {MIT press Cambridge}.

\bibitem[{Sutton et~al.(2009)Sutton, Maei, Precup, Bhatnagar, Silver,
  Szepesv{\'a}ri and Wiewiora}]{sutton2009fast}
Sutton RS, Maei HR, Precup D, Bhatnagar S, Silver D, Szepesv{\'a}ri C and
  Wiewiora E (2009) Fast gradient-descent methods for temporal-difference
  learning with linear function approximation.
\newblock In: \emph{Proceedings of the 26th Annual International Conference on
  Machine Learning}. pp. 993--1000.

\bibitem[{Sutton et~al.(2011)Sutton, Modayil, Delp, Degris, Pilarski, White and
  Precup}]{sutton_horde_2011}
Sutton RS, Modayil J, Delp M, Degris T, Pilarski PM, White A and Precup D
  (2011) Horde: {{A}} scalable real-time architecture for learning knowledge
  from unsupervised sensorimotor interaction.
\newblock In: \emph{{{AAMAS}} 2011}. {International Foundation for Autonomous
  Agents and Multiagent Systems}, pp. 761--768.

\bibitem[{Sutton et~al.(1999)Sutton, Precup and Singh}]{sutton_between_1998}
Sutton RS, Precup D and Singh S (1999) Between mdps and semi-mdps: A framework
  for temporal abstraction in reinforcement learning.
\newblock \emph{Artificial Intelligence} 112(1): 181 -- 211.

\bibitem[{White(2015)}]{white_developing_2015}
White A (2015) \emph{Developing a Predictive Approach to Knowledge}.
\newblock {{PhD Thesis}}, University of Alberta.

\bibitem[{Wolpert et~al.(1995)Wolpert, Ghahramani and
  Jordan}]{wolpert_internal_1995}
Wolpert DM, Ghahramani Z and Jordan MI (1995) An internal model for
  sensorimotor integration.
\newblock \emph{Science} 269(5232): 1880--1882.

\end{thebibliography}
\end{document}